\documentclass[letterpaper,10pt,conference]{ieeeconf}

\IEEEoverridecommandlockouts
\usepackage{amsmath}
\usepackage{amssymb}
\usepackage{graphicx}
\usepackage{booktabs}
\usepackage{siunitx}
\usepackage{cite}
\usepackage{url}

\newcommand{\method}{PreferenceFlow}

\title{\LARGE \bf
PreferenceFlow: Test-Time Guidance of Flow-Matching\\
Robot Policies from Human Interventions
}

\author{Yiqi Tang$^{1}$, Diyuan Shi$^{1}$, Runze Li$^{1}$, Donglin Wang$^{1,*}$%
\thanks{$^{1}$All authors are with Westlake University.}%
\thanks{$^{*}$Corresponding Author.}%
}

\begin{document}
\maketitle
\thispagestyle{empty}
\pagestyle{empty}

\begin{abstract}
Flow-matching policies can represent complex robot behaviors but remain
susceptible to local errors under distribution shift at deployment. Many
reinforcement learning approaches to policy improvement require reward
signals that are difficult to specify or obtain in real-world manipulation.
We present \method{}, a framework for improving a pretrained flow policy at
test time without environment rewards or updates to the base policy. Human
intervention chunks are paired with robot chunks generated from the same
initial conditioning state to train a preference model. During inference,
we adopt the QGF sampling update, replacing its value gradient with the
preference gradient evaluated at an estimated clean action. A gradient-cap
loss penalizes excessive gradients on intervention pairs, while a
zero-gradient loss discourages guidance near actions from expert
demonstrations. On four real-world precision insertion tasks with a Franka
robot, \method{} achieves a mean success rate of 90.5\%, compared with 69\%
for the frozen policy. Ablations support the roles of gradient regularization
and correctly ordered preference labels in the evaluated settings. These
results demonstrate the utility of human interventions as local preference
supervision for guiding frozen generative robot policies.

\end{abstract}

\section{Introduction}
\label{sec:introduction}

Robot foundation models aim to support general-purpose control across tasks.
Recent policies combine pretrained vision-language representations with
diffusion or flow-matching action decoders to model multimodal distributions
over continuous action chunks~\cite{chi2023diffusion,lipman2023flow,
black2024pi0,black2025pi05}. Despite their broad capabilities, these policies
remain vulnerable to local execution errors. Changes in visual appearance,
unfamiliar object configurations, and contact-rich interactions can require
corrections even when the policy possesses the skills needed to complete a task.

Deployment experience provides an opportunity to address these errors.
However, updating all parameters of a robot foundation model is computationally
expensive and may affect capabilities acquired during pretraining. Test-time
guidance instead modifies sampling through the gradient of an auxiliary
objective while keeping the base policy fixed. Q-Guided Flow (QGF), for
example, uses a learned critic's action gradient to favor higher-value
actions~\cite{zhou2026qgf}. Although this approach avoids actor updates, it
requires an informative Q-function. On real robots, rewards may be sparse,
delayed, or unavailable, and critic gradients at actions outside the training
distribution may be poorly constrained.

Reinforcement learning from human feedback (RLHF) for language
models~\cite{ouyang2022training} suggests an alternative source of supervision:
relative judgments can be easier to provide than explicit reward functions.
In robot manipulation, an operator can indicate a corrective action by
intervening during execution. We interpret the resulting human action chunk
as preferred to a robot candidate regenerated from the same window-start
state. These comparisons provide local supervision at states where the
policy requires correction. We therefore investigate whether an
intervention-trained preference model can guide a pretrained flow policy
without learning a Q-function or updating the base policy.

We propose \method{}, a framework that learns a state-conditioned preference
score from human--robot action-chunk comparisons using a Bradley--Terry
objective. A KL-regularized objective motivates an ideal distribution that
exponentially reweights the base policy by this score. For practical
inference, we adopt QGF's flow-guidance update~\cite{zhou2026qgf}, replacing
its Q-function with the learned preference score. At each sampling step, we
evaluate the preference gradient at an estimated clean action and add it to
the frozen velocity field.

\begin{figure*}[t]
    \centering
    \includegraphics[width=0.98\textwidth]%
        {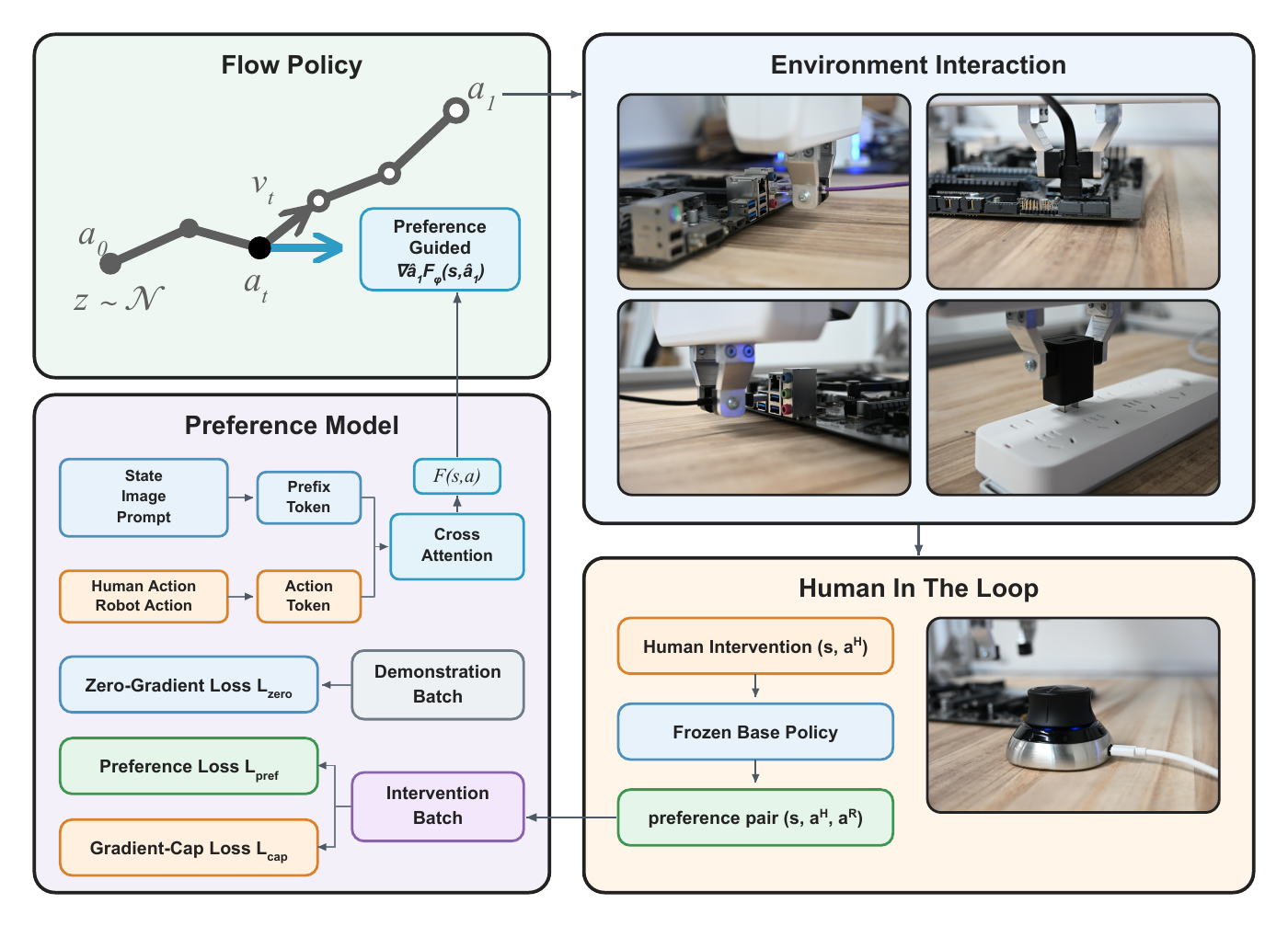}
    \caption{Overview of \method{}. Human interventions provide action-chunk
    comparisons sharing a window-start state, while expert demonstrations
    provide anchors for zero-gradient regularization. Separate intervention
    and demonstration sub-batches train a state-conditioned preference model.
    Its action gradient guides the frozen $\pi_{0.5}$ flow sampler at inference.
    Only the preference-model parameters are updated during adaptation.}
    \label{fig:overview}
\end{figure*}

Our contributions are threefold:
\begin{itemize}
    \item A framework for learning state-conditioned action preferences from
    human interventions. It constructs human--robot comparisons sharing an
    initial state and scores actions using frozen policy conditioning features.
    The learned score guides the existing QGF update without environment
    rewards or base-policy updates.
    \item Two complementary gradient regularizers for preference-based
    guidance. A gradient-cap loss penalizes excessive gradients on intervention
    pairs, while a zero-gradient loss suppresses guidance near demonstrated
    routine actions.
    \item An evaluation on four real-world precision insertion tasks, where
    \method{} achieves 90.5\% mean success, exceeding the frozen policy by
    21.5 percentage points. Ablations examine gradient regularization, model
    architecture, intervention-data quantity, and preference-label ordering.
\end{itemize}

\section{Related Work}
\label{sec:related}

\subsection{Generative Robot Policies and Robot Foundation Models}

\begingroup
\setlength{\parskip}{0pt}
Generative models represent multimodal distributions over robot actions. Diffusion policies generate action sequences by
iteratively denoising samples conditioned on observations, enabling
visuomotor control with temporally coordinated action
chunks~\cite{chi2023diffusion}. Flow matching instead learns a velocity field
that transports noise to actions~\cite{lipman2023flow}; robot policies such as
$\pi_0$ use this formulation to generate continuous action chunks conditioned
on visual and language inputs~\cite{black2024pi0}. Both formulations support expressive robot policies whose iterative
sampling procedures can be modified through test-time guidance.

Modern robot foundation models combine action generation with large-scale
pretraining to support generalization across tasks and environments.
Vision-language-action (VLA) models integrate visual and linguistic knowledge
with robot control~\cite{black2025pi05,gemini2025robotics15,nvidia2025groot,shukor2025smolvla}.
World-action models (WAMs) further couple future visual-state prediction with
executable action generation, connecting predictive world modeling to
control~\cite{ye2026dreamzero,kim2026cosmos,yan2026flexpi,wang2026openwam,hung2026modar}.
These models provide capable policy priors, but local deployment errors
motivate further adaptation. We study preference-based guidance of a frozen
flow policy, using $\pi_{0.5}$ as the experimental backbone.
\par
\endgroup

\subsection{Improving Pretrained Robot Policies}

Pretrained robot policies can be improved by updating policy parameters,
learning auxiliary controllers, or modifying action generation at inference.
For parameter adaptation, $\pi^{*}_{0.6}$ uses RECAP, an iterative offline RL procedure
that combines demonstrations, autonomous rollouts, and human interventions;
a learned value function supplies reward-derived advantages for
advantage-conditioned VLA training~\cite{physicalintelligence2025pistar06}.
Other approaches learn auxiliary modules: DSRL trains a policy in the latent
noise space of a frozen diffusion policy~\cite{wagenmaker2025dsrl}, while
RL Token trains small actor--critic heads on a compact representation from
a pretrained robot foundation model~\cite{xu2026rltoken}. QGF instead uses
a learned critic's action gradient to guide flow sampling while keeping the
base policy frozen~\cite{zhou2026qgf}. These methods use reward or value information to improve policies through
distinct mechanisms.

Other methods incorporate external action guidance or runtime constraints. UniSteer, Flow Reversal Steering (FRS), and
FlowDAgger map external action guidance back to initial
noise~\cite{lu2026unisteer,tang2026frs,flowdagger}. UniSteer combines inverted
human corrections with noise-space reinforcement learning; FRS projects
coarse guidance onto the generalist policy's action modes; FlowDAgger trains
a latent policy from inverted interventions. Lexicographic steering uses
prioritized runtime costs and candidate filtering to steer frozen diffusion
and flow policies~\cite{jia2026lexicographic}. Our method learns a
differentiable local preference objective from human--robot action-chunk
comparisons sharing an initial state, then uses its action gradient to improve
a frozen flow policy without environment rewards or base-policy updates.

\subsection{Human Preferences and Interactive Robot Learning}

Pairwise human judgments provide supervision for preference-based reinforcement
learning~\cite{christiano2017deep} and RLHF for language models~\cite{ouyang2022training}.
T-REX learns rewards from ranked demonstrations~\cite{brown2019trex}, while
PEBBLE improves feedback efficiency through preference queries and experience
relabeling~\cite{lee2021pebble}. Direct Preference Optimization (DPO) uses the
closed-form solution of a KL-regularized reward objective to optimize a policy
directly from comparisons~\cite{rafailov2023dpo}. We use the same exponential
tilting principle to motivate inference-time guidance, while keeping the
robot policy parameters frozen.

Interactive imitation learning collects expert supervision at states visited
by the learner. DAgger aggregates expert labels~\cite{ross2011dagger};
SafeDAgger reduces expert queries through a learned safety policy
~\cite{zhang2017safedagger}; HG-DAgger allows humans to control takeovers
~\cite{kelly2019hgdagger}. HIL-SERL combines demonstrations, human interventions,
and reward-based online learning for dexterous manipulation~\cite{luo2024hilserl}.
In our setting, intervention-derived action-chunk comparisons train a
preference model whose action gradient guides a frozen flow sampler, without
environment rewards or base-policy updates.

\section{Problem Formulation}
\label{sec:problem}

Let \(s\in\mathcal{S}\) denote the policy's conditioning input and
\(a\in\mathbb{R}^{H\times D}\) an action chunk with horizon \(H\) and action
dimension \(D\). A pretrained flow policy \(\hat\pi\) generates chunks
conditioned on \(s\). Flow-matching training uses the linear conditional path
\(x_t=(1-t)z+ta\), where \(z\sim\mathcal{N}(0,I)\) and
\(t\in[0,1]\) increases from noise to action. At inference, \(a_t\) denotes
the evolving sample, whose trajectory need not follow an individual linear
training path. Starting from \(a_0\sim\mathcal{N}(0,I)\), the velocity field
\(v_\theta(s,a_t,t)\) generates actions using \(T\) Euler steps with
\(\delta=1/T\):
\begin{equation}
    a_{t+\delta}=a_t+\delta v_\theta(s,a_t,t),
    \quad t=0,\delta,\ldots,1-\delta.
    \label{eq:base-sampler}
\end{equation}
The pretrained parameters \(\theta\) remain frozen throughout adaptation.

Consider an intervention trajectory
\((s_0,u_0^H,s_1,u_1^H,\ldots)\), where \(u_i^H\) is a single human action. For each complete length-\(K\) window
beginning at \(s_i\), we define the human action chunk
\(a_i^H=(u_i^H,\ldots,u_{i+K-1}^H)\). We sample a full robot chunk
\(\tilde a_i^R\sim\hat\pi(\cdot\mid s_i)\) and retain its first \(K\leq H\)
steps, \(a_i^R=\tilde a_i^{R,[0:K-1]}\), for comparison. Thus, the two chunks
share the same initial conditioning state \(s_i\). We assign the preference label
\(a_i^H\succ a_i^R\) for intervention-derived pairs. Subsequent human actions are collected under closed-loop execution. Thus,
shared initial conditioning does not imply identical state trajectories for
the two action chunks.

The resulting preference dataset is
\begin{equation}
    \mathcal{D}_{\mathrm{pref}}
    =\{(s_i,a_i^H,a_i^R)\}_{i=1}^{N}
    \label{eq:dataset}
\end{equation}
which we use to learn local action preferences for guiding the frozen policy,
without learning an environment reward or Q-function.

\section{Preference-Guided Flow Policies}
\label{sec:method}

Fig.~\ref{fig:overview} summarizes preference-data construction, preference
learning, and guided flow inference in \method{}.

\subsection{Preference Model Architecture}
\label{sec:pref-architecture}

\begin{figure}[t]
    \centering
    \includegraphics[width=0.98\columnwidth,trim=0 100bp 0 0,clip]%
        {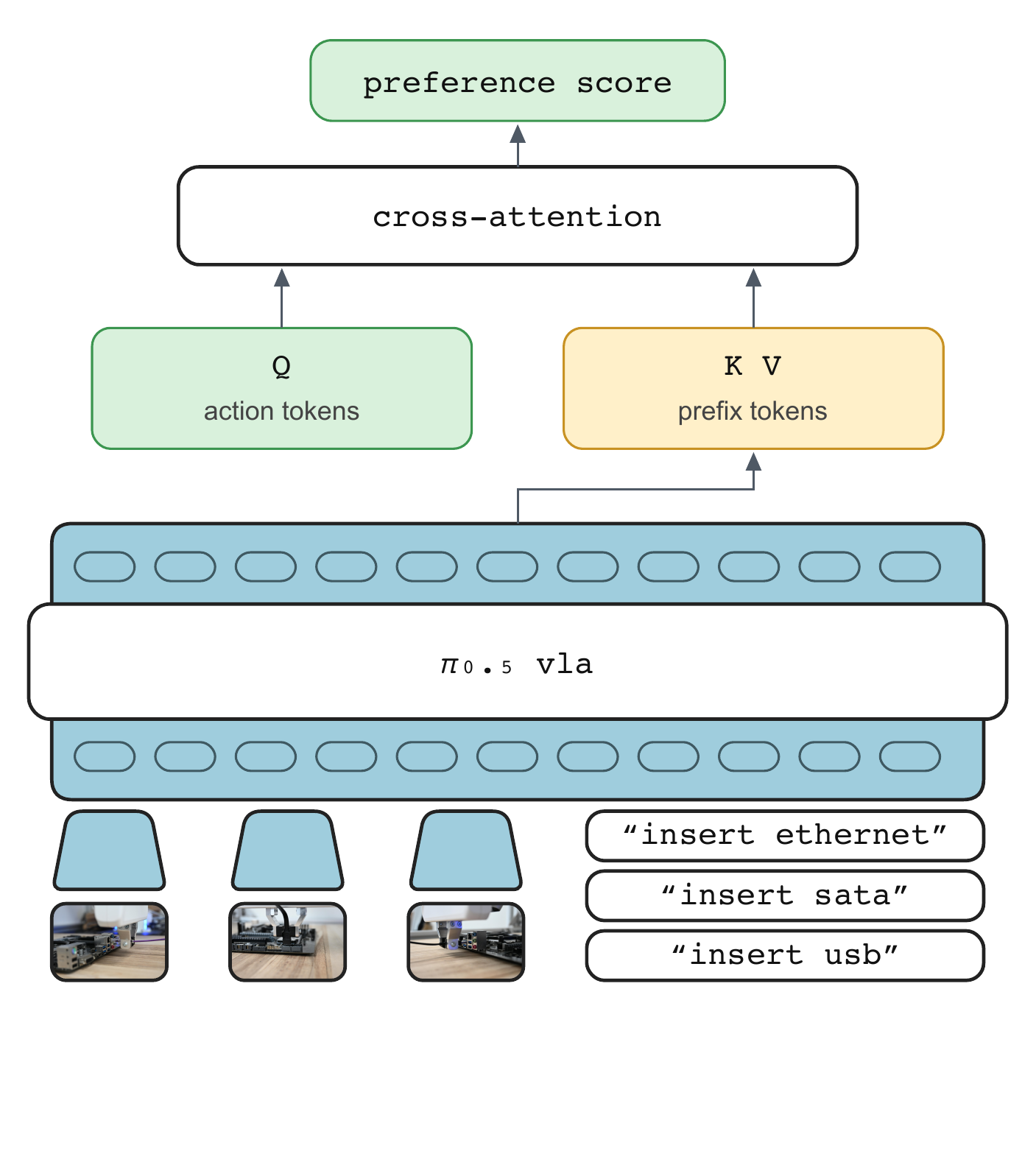}
    \caption{State-conditioned preference-model architecture. Three camera
    images, the language instruction, and robot-state information form the
    frozen $\pi_{0.5}$ conditioning representation, used as keys and values. Candidate action
    tokens form the queries and cross-attend to the prefix tokens to produce
    the preference score \(F_\phi(s,a)\). Its action gradient provides the
    guidance signal for flow sampling. The schematic omits robot-state inputs
    and the per-step score heads followed by mean aggregation.}
    \label{fig:preference-model}
\end{figure}

We use $\pi_{0.5}$~\cite{black2025pi05} as the frozen flow policy and score the
first \(K\leq H\) steps of each action chunk. The preference model reuses the
policy's frozen conditioning representation
(Fig.~\ref{fig:preference-model}). Let
\(Z_s\in\mathbb{R}^{P\times d_p}\) denote the valid visual, language, and
robot-state tokens for state \(s\), and let \(m_s\) be their attention mask.
A learned projection maps these tokens to width \(d\). Each candidate action
step \(a^{[j]}\) is projected into an action token with a temporal position
embedding. These tokens independently cross-attend to \(Z_s\), without
temporal self-attention or communication between action steps. A score head
then maps each resulting state-conditioned representation to a scalar and
averages the per-step scores:
\begin{equation}
    \begin{aligned}
    c_j &= \operatorname{CrossAttn}_\phi(q_j(a^{[j]}),Z_s,m_s),\\
    f_j &= h_\phi(c_j,j),\\
    F_\phi(s,a^{[0:K-1]}) &= \frac{1}{K}\sum_{j=0}^{K-1} f_j.
    \end{aligned}
    \label{eq:preference-model}
\end{equation}
The score head receives only the cross-attention outputs; there is no direct
action-only residual from \(q_j(a^{[j]})\) to \(f_j\). This bottleneck conditions each score on the state while retaining a separate
gradient for each action step. Independent scoring is intended to limit
reliance on global action-style cues.

\subsection{Preference Model Training}
\label{sec:pref-training}

We train the model from intervention-derived pairs with a Bradley--Terry
likelihood~\cite{christiano2017deep}:
\begin{equation}
    p_\phi(a^H\succ a^R\mid s)=\sigma\!\left(
    \frac{F_\phi(s,a^H)-F_\phi(s,a^R)}{\tau}\right),
    \label{eq:bt-probability}
\end{equation}
where \(\sigma\) is the logistic sigmoid and \(\tau>0\) is a temperature. The loss is
\begin{equation}
    \mathcal{L}_{\mathrm{pref}}(\phi)
    =-\mathbb{E}_{(s,a^H,a^R)\sim\mathcal{D}_{\mathrm{pref}}}
    \log p_\phi(a^H\succ a^R\mid s).
    \label{eq:pref-loss}
\end{equation}
The pairwise likelihood depends on score differences; it supervises relative
action preferences without calibrating the score to an expected return.

The ranking objective does not explicitly regularize the action gradients
used during sampling or discourage guidance near routine expert behavior. We therefore introduce two complementary gradient objectives. Let
\(\mathcal{D}_{\mathrm{int}}=\mathcal{D}_{\mathrm{pref}}\) denote intervention pairs and
\(\mathcal{D}_{\mathrm{demo}}\) contain states and actions from complete expert
demonstrations. On intervention data, we retain the preference loss and add a
soft gradient-cap penalty,
\begin{equation}
    \begin{aligned}
    \mathcal{L}_{\mathrm{cap}}
    &=\mathbb{E}_{\mathcal{D}_{\mathrm{int}}}
    \left[\frac{\ell_{\mathrm{cap}}(s,a^H)+\ell_{\mathrm{cap}}(s,a^R)}{2}\right],\\
    \ell_{\mathrm{cap}}(s,a)
    &=\left[\max\!\left(0,\|\nabla_aF_\phi(s,a)\|_2-g_{\max}\right)\right]^2.
    \end{aligned}
    \label{eq:gradient-cap}
\end{equation}
The expectation is over pairs \((s,a^H,a^R)\), with their two penalties
averaged. All actions are normalized, and gradients are evaluated in these
coordinates. The penalty acts only above \(g_{\max}\), discouraging excessive
gradients while retaining smaller corrective gradients.

Expert demonstrations provide anchors for suppressing unnecessary guidance.
For a demonstrated action chunk \(a^E\), we penalize preference gradients
in its local neighborhood:
\begin{equation}
    \begin{aligned}
    \mathcal{L}_{\mathrm{zero}}(\phi)
    &=\mathbb{E}_{\mathcal{D}_{\mathrm{demo}},\,\epsilon}
    \left[\|\nabla_aF_\phi(s,a^E+\epsilon)\|_2^2\right],\\[-1mm]
    &\qquad \epsilon\sim\mathcal{N}(0,\sigma_a^2I).
    \end{aligned}
    \label{eq:zero-gradient}
\end{equation}
Combining preference supervision with these two regularizers gives
\begin{equation}
    \mathcal{L}_{\mathrm{total}}
    =\mathcal{L}_{\mathrm{pref}}
    +\lambda_{\mathrm{cap}}\mathcal{L}_{\mathrm{cap}}
    +\lambda_{\mathrm{zero}}\mathcal{L}_{\mathrm{zero}}.
    \label{eq:total-pref-loss}
\end{equation}
Here, \(\lambda_{\mathrm{cap}}\) and \(\lambda_{\mathrm{zero}}\) weight the
two regularizers. Intervention pairs supply the preference and gradient-cap
losses, while expert demonstrations supply the zero-gradient loss. The combined objective encourages corrective gradients at intervention
states and small gradients near demonstrated actions. Only the
preference-model parameters \(\phi\) are optimized; the base policy and its
conditioning encoder remain frozen.

\begin{figure*}[t]
    \centering
    \includegraphics[width=0.98\textwidth]{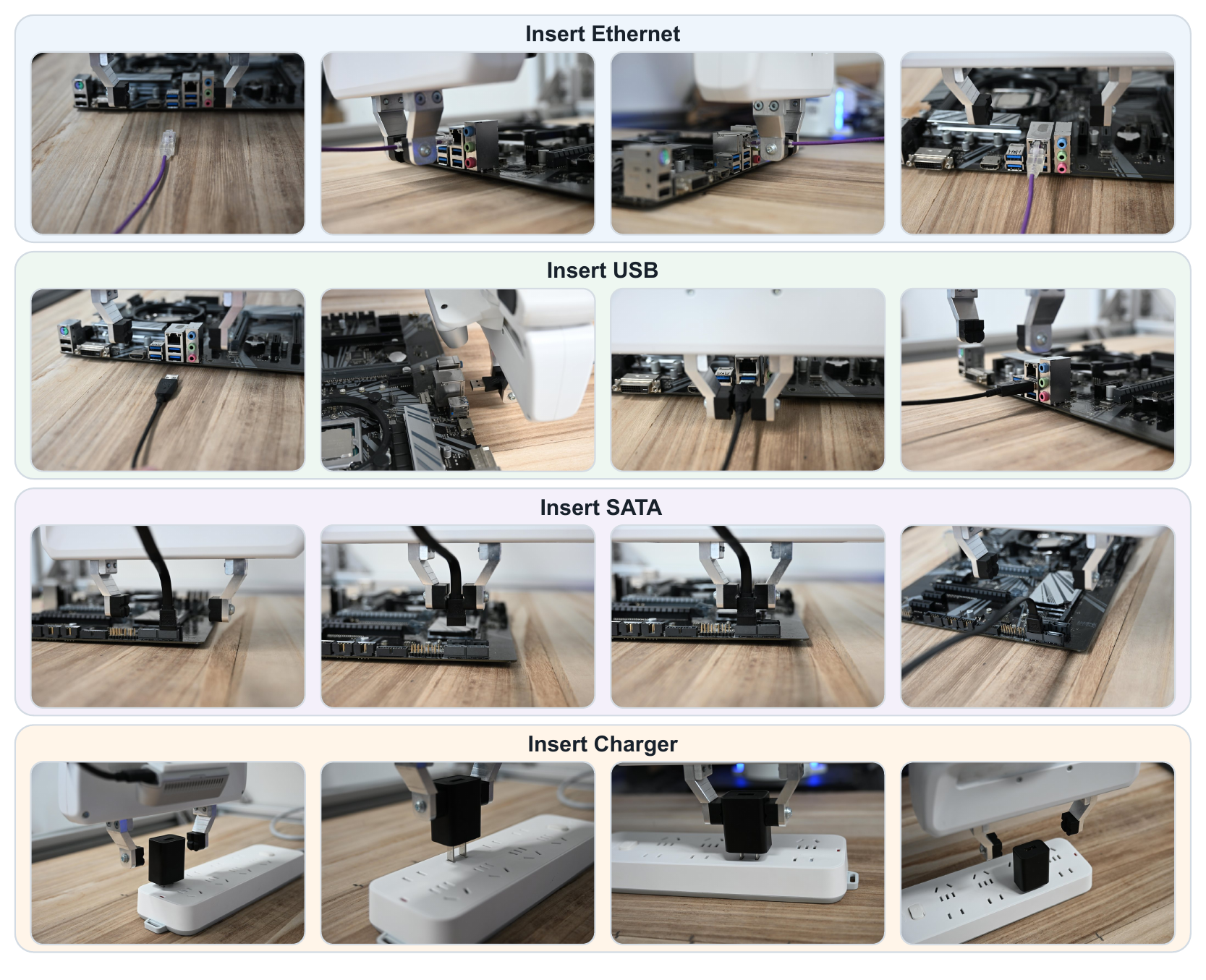}
    \caption{Real-world precision insertion tasks. Each row shows one task and
    progresses from left to right, from the initial approach to insertion.}
    \label{fig:task-progress}
\end{figure*}

\subsection{Preference-Guided Flow Sampling}
\label{sec:preference-objective}

The preference score motivates a policy that favors preferred actions while
remaining close to the frozen policy \(\hat\pi\). At a fixed state \(s\),
consider the KL-regularized objective
\begin{equation}
    \begin{aligned}
    \mathcal{J}_s(\pi)
    & =
    \mathbb{E}_{a\sim\pi(\cdot\mid s)}[F_\phi(s,a)]
    \\
    &\quad-\beta D_{\mathrm{KL}}\!\left(
        \pi(\cdot\mid s)\,\|\,\hat\pi(\cdot\mid s)\right),
    \end{aligned}
    \label{eq:regularized-objective}
\end{equation}
where \(\beta>0\) controls deviation from the base distribution. Maximizing
over normalized densities supported by \(\hat\pi\) gives
the exponential tilt used in KL-regularized preference optimization~\cite{rafailov2023dpo}:
\begin{equation}
    \begin{aligned}
    \pi_\phi^\star(a\mid s)
        &=\frac{\hat\pi(a\mid s)}{Z_\phi(s)}
          \exp\!\left(\frac{F_\phi(s,a)}{\beta}\right),\\
    Z_\phi(s)&=\int\hat\pi(a\mid s)
          \exp\!\left(\frac{F_\phi(s,a)}{\beta}\right)\mathrm{d}a.
    \end{aligned}
    \label{eq:preference-tilted-policy}
\end{equation}
where the partition function \(Z_\phi(s)\) is assumed finite. This distribution
assigns greater relative probability to higher-scoring actions within the
base policy's support. It motivates preference-based improvement but does
not determine a flow velocity field.

For practical inference, we adopt QGF's flow-guidance
update~\cite{zhou2026qgf}, replacing its learned Q-function with the preference
score \(F_\phi\). Because the preference model is trained on clean actions,
we first estimate the clean endpoint from the current noisy action using
one Euler extrapolation:
\begin{equation}
    \hat a_1=a_t+(1-t)v_\theta(s,a_t,t).
    \label{eq:clean-action}
\end{equation}
We evaluate the preference gradient with respect to this estimated clean
action, without differentiating through the frozen velocity field. Following
QGF, the guided Euler update is
\begin{equation}
    a_{t+\delta}=a_t+\delta\left(
    v_\theta(s,a_t,t)+\lambda_{\mathrm{guide}}
    \nabla_{\hat a_1}F_\phi(s,\hat a_1)\right),
    \label{eq:guided-sampler}
\end{equation}
with \(\delta=1/T\) and a tunable guidance weight
\(\lambda_{\mathrm{guide}}\geq0\). Here, \(F_\phi(s,\hat a_1)\) evaluates only
the first \(K\) steps of the full estimated chunk; its direct gradient for
the remaining steps is zero. The guidance weight is selected empirically, independently of the conceptual
KL coefficient \(\beta\). We recompute the gradient at every sampling step;
setting \(\lambda_{\mathrm{guide}}=0\) recovers the base sampler. Both models
remain fixed during inference. This adaptation supplies a learned preference
objective to QGF's existing update and does not guarantee exact sampling from
Eq.~\eqref{eq:preference-tilted-policy}.


\section{Experimental Evaluation}
\label{sec:experiments}

Our experiments address four questions: (1) Does \method{} improve
real-robot task success over the frozen base policy, and how does its observed
performance compare with intervention-, value-, and inversion-based baselines?
(2) How do the gradient-cap and zero-gradient objectives affect performance
at a shared guidance weight and after adjustment for stable deployment?
(3) How do the amount of intervention data and the
preference-model architecture affect task success? (4) How does randomizing
preference supervision affect task success, and can the learned score also
support sample-based action reranking?

\subsection{Tasks and Experimental Setup}

We evaluate a single Franka robot on four precision insertion tasks:
\textit{Insert Ethernet}, \textit{Insert USB}, \textit{Insert SATA}, and
\textit{Insert Charger}. These tasks require visual localization of connectors
and ports, followed by precise position and orientation alignment under tight
geometric tolerances. Small localization or pose errors can prevent insertion,
making local corrections essential. The setup includes one wrist-mounted
camera and two fixed cameras, with a control frequency of 30~Hz. Each method
is evaluated in 50 trials per task; success rate is the fraction of successful
trials.

\subsection{Data and Implementation Details}

\begingroup
\setlength{\parskip}{0pt}

Models are trained separately for each task. We construct preference pairs
from complete intervention windows of \(K=10\) steps with stride one. Each
human chunk is paired with the first 10 steps of a full robot chunk sampled
from the frozen policy at the same window-start state. Data splits are grouped
by rollout. During deployment, the robot executes the first 10 actions of each
chunk before replanning from a new observation.

\begin{table*}[!t]
    \centering
    \small
    \caption{Task success rates (\%) over 50 real-robot trials per task.
    Mean denotes the arithmetic mean across the four tasks. $\Delta$ denotes the absolute
    improvement over the frozen policy in percentage points (pp).
    Bold indicates the best observed result in each column.}
    \label{tab:main-results}
    \begin{tabular}{l*{4}{S[table-format=2.0]}S[table-format=2.1]S[table-format=+2.1,retain-explicit-plus=true]}
        \toprule
        Method & {Ethernet} & {USB} & {SATA} & {Charger}
        & {Mean $\uparrow$} & {$\Delta$ vs.\ base (pp) $\uparrow$} \\
        \midrule
        Frozen $\pi_{0.5}$ & 66 & 70 & 68 & 72 & 69.0 & \multicolumn{1}{c}{---} \\
        Intervention BC & 82 & 84 & 88 & 84 & 84.5 & +15.5 \\
        QGF & 74 & 74 & 80 & 76 & 76.0 & +7.0 \\
        FlowDAgger & 78 & 80 & 76 & 80 & 78.5 & +9.5 \\
        \textbf{PreferenceFlow (ours)} & \bfseries 90 & \bfseries 92 & \bfseries 92
        & \bfseries 88 & \bfseries 90.5 & \bfseries +21.5 \\
        \bottomrule
    \end{tabular}
\end{table*}

For each task, we first fine-tune all $\pi_{0.5}$ parameters by supervised
learning on 50 expert demonstrations. The resulting checkpoint remains frozen
during preference learning and guided inference. Each preference-model update
uses two sub-batches: intervention pairs supervise preference ranking and
gradient-cap regularization, while states and actions sampled from the 50
complete demonstrations supervise the zero-gradient objective. The losses
are combined according to Eq.~\eqref{eq:total-pref-loss}. QGF, FlowDAgger,
and PreferenceFlow are each trained on a single RTX~5090 GPU.

\par
\endgroup

\subsection{Baselines}

We compare against three adaptation methods and the frozen base policy.
\textbf{Frozen base policy} is the task-adapted $\pi_{0.5}$ checkpoint described
above. \textbf{Intervention BC} initializes from this checkpoint and fine-tunes
it on human takeover windows using the human actions as regression targets.
\textbf{QGF} uses the authors' official code~\cite{zhou2026qgf}. Because that
implementation does not provide a $\pi_{0.5}$ configuration, we use the same
state-conditioned network architecture as our preference model for its
Q-function. The critic is trained on 50 failed base-policy rollouts and the
50 successful expert demonstrations, using binary trajectory-outcome labels,
and remains fixed during inference. \textbf{FlowDAgger} uses the official
$\pi_{0.5}$ implementation to learn structured initial noise from 50
human-intervention episodes per task~\cite{flowdagger}.

\textbf{PreferenceFlow} trains its preference model offline on the same 50
human-intervention episodes per task. An episode may contain multiple takeover
segments, with a cumulative intervention duration of approximately 2--5 seconds.
The number and duration of individual segments vary across episodes.

\subsection{Main Results}

Table~\ref{tab:main-results} summarizes the main comparison. PreferenceFlow
achieves the highest observed success rate among these methods on all four
tasks, with a mean of \(90.5\%\), compared with \(69\%\) for the frozen policy.
This corresponds to 181 versus 138 successes in 200 trials, an absolute gain
of 21.5 percentage points. The gains on \textit{Insert Ethernet},
\textit{Insert USB}, \textit{Insert SATA}, and \textit{Insert Charger} are
24, 22, 24, and 16 percentage points, respectively.

Intervention BC achieves \(84.5\%\) mean success, the highest among the three
adaptation baselines. PreferenceFlow exceeds this result by 6 percentage
points while keeping the base policy frozen, with gains of 8, 8, 4, and 4
percentage points on the four tasks. These results support preference guidance
as an alternative to full-parameter fine-tuning on interventions.

QGF achieves \(76\%\) mean success, improving all four tasks and exceeding the
frozen policy by 7 percentage points. PreferenceFlow performs 14.5 percentage
points better on average. One possible explanation is that binary trajectory
outcomes provide limited supervision for local corrective actions, whereas
intervention pairs directly supervise local action preferences.

FlowDAgger achieves \(78.5\%\) mean success, improving all four tasks and
exceeding the frozen policy by 9.5 percentage points. PreferenceFlow exceeds
FlowDAgger by 12 percentage points. In the evaluated configuration, FlowDAgger
inverts complete 50-step chunks and therefore cannot directly use shorter
takeover segments, even when cumulative intervention time spans more than one
chunk. PreferenceFlow uses complete 10-step windows from each segment. This
difference may affect the amount of usable supervision; the present comparison
does not isolate its contribution to the performance gap.

\subsection{Gradient Objective Ablation}
\label{sec:ablations-gradient}

We compare the full objective with three variants: removing the gradient-cap
loss, removing the zero-gradient loss, and retaining only the preference loss.
All variants use the same intervention pairs. Removing the zero-gradient loss
also removes the demonstration-based anchors from preference-model training;
this ablation therefore evaluates the contribution of that supervision together
with its regularization objective.

\begin{table}[t]
    \centering
    \small
    \caption{Gradient-objective ablation: mean success (\%) across four tasks.
    Fixed guidance uses the full model's weight. For adjusted guidance, each
    ablated variant's weight is progressively reduced until deployment is stable.
    Unstable indicates that stable deployment was not possible at the tested
    weight; no success rate is assigned. Bold indicates the best observed rates.}
    \label{tab:gradient-objective-ablation}
    \setlength{\tabcolsep}{3pt}
    \begin{tabular}{lS[table-format=2.1]S[table-format=2.1]}
        \toprule
        Training objective & \multicolumn{2}{c}{Mean success (\%) $\uparrow$} \\
        \cmidrule(lr){2-3}
        & {\shortstack{Fixed\\guidance}} & {\shortstack{Adjusted\\guidance}} \\
        \midrule
        Full objective & \bfseries 90.5 & \bfseries 90.5 \\
        Without gradient-cap loss & 15.0 & 81.0 \\
        Without zero-gradient loss & 5.0 & 74.5 \\
        Preference loss only & \multicolumn{1}{c}{Unstable} & 69.0 \\
        \bottomrule
    \end{tabular}
\end{table}

Table~\ref{tab:gradient-objective-ablation} reports two evaluation settings.
At the full model's guidance weight, removing the gradient-cap or zero-gradient
loss reduces mean success from \(90.5\%\) to \(15\%\) or \(5\%\), respectively.
The preference-only variant cannot be deployed stably at this weight, so no
success rate is reported. These results indicate substantial degradation
when either regularizer is removed at the tested guidance strength.

We then progressively reduce each ablated variant's guidance weight until
stable deployment is possible. Mean success reaches \(81\%\) without the
gradient-cap loss and \(74.5\%\) without the zero-gradient loss, remaining
9.5 and 16 percentage points below the full objective. The preference-only
variant achieves \(69\%\), matching the frozen-policy mean and remaining
21.5 percentage points below the full objective. Both evaluation settings
support the regularizers under the tested deployment conditions; the adjusted
weights do not recover the full model's observed performance.

\subsection{Human-Intervention Data Ablation}
\label{sec:ablations-intervention-data}

\begin{figure}[t]
    \centering
    \includegraphics[width=\columnwidth]{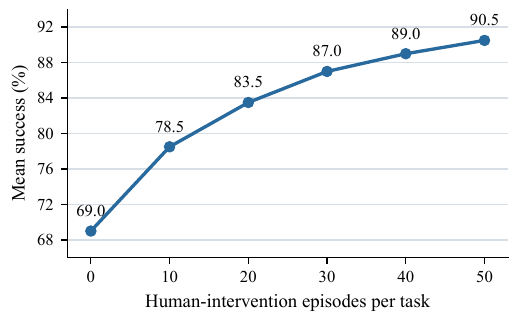}
    \caption{Mean success (\%) across four tasks as a function of intervention
    episodes per task. The zero-episode condition is the frozen policy without
    preference guidance.}
    \label{fig:intervention-data-ablation}
\end{figure}

We vary the number of intervention episodes per task while keeping the expert
demonstrations and training settings fixed. The zero-episode condition in
Fig.~\ref{fig:intervention-data-ablation} is the frozen policy without guidance,
which achieves \(69\%\) mean success. With 10, 20, 30, 40, and 50 intervention
episodes, mean success increases to \(78.5\%\), \(83.5\%\), \(87\%\), \(89\%\),
and \(90.5\%\), respectively. Ten episodes already provide a gain of 9.5
percentage points over the frozen policy. Each subsequent block of 10 episodes
adds 5, 3.5, 2, and 1.5 percentage points, respectively. The observed gains
thus diminish with additional data but remain positive over the tested range.

\subsection{Preference Model Architecture Ablation}
\label{sec:ablations-architecture}

Adding temporal self-attention among action tokens before score prediction
reduces mean success from \(90.5\%\) to \(77\%\), a difference of 13.5 percentage
points (Table~\ref{tab:architecture-ablation}). This result supports independent
action-token scoring for the evaluated tasks. Temporal self-attention may
allow the scorer to exploit global patterns associated with human control
style rather than task-relevant preferences. This interpretation remains a
hypothesis: the ablation compares architectures but does not establish the
mechanism underlying the performance difference.

\begin{table}[t]
    \centering
    \small
    \caption{Preference-model architecture ablation: mean success (\%) across
    four tasks. Both architectures use state-conditioned cross-attention;
    the ablated variant additionally uses temporal self-attention among action
    tokens. Bold indicates the best observed rate.}
    \label{tab:architecture-ablation}
    \setlength{\tabcolsep}{3pt}
    \begin{tabular}{lS[table-format=2.1]}
        \toprule
        Architecture & {\shortstack{Mean success\\(\%) $\uparrow$}} \\
        \midrule
        \shortstack[l]{Cross-attention +\\temporal self-attention} & 77.0 \\
        Per-step cross-attention (ours) & \bfseries 90.5 \\
        \bottomrule
    \end{tabular}
\end{table}

\subsection{Score-Based and Negative Controls}
\label{sec:ablations-controls}

Preference reranking samples five chunks from the frozen $\pi_{0.5}$ policy
at each decision and selects the candidate with the highest preference score.
It achieves \(92\%\) mean success, compared with \(90.5\%\) for gradient guidance
(Table~\ref{tab:score-controls}), a difference of 1.5 percentage points. This
result supports the score's utility for selecting among base-policy candidates.
Reranking requires five complete samples per decision, whereas PreferenceFlow
guides one sample with preference-gradient evaluations at each flow step.
Their relative latency also depends on the cost of these gradient evaluations.

For the negative control, we randomly reverse the preference direction within
human--robot pairs, assigning the robot action as preferred in flipped pairs.
Mean success decreases to \(63.5\%\), which is 27 percentage points below
PreferenceFlow and 5.5 percentage points below the frozen policy. Randomized
supervision can therefore degrade observed performance below the base-policy
level. Together, these controls support the utility of the learned scores
and the importance of preference-label ordering.

\begin{table}[t]
    \centering
    \small
    \caption{Score-based selection and negative controls: mean success (\%)
    across four tasks. Bold indicates the best observed rate.}
    \label{tab:score-controls}
    \setlength{\tabcolsep}{3pt}
    \begin{tabular}{lS[table-format=2.1]}
        \toprule
        Variant & {Mean success (\%) $\uparrow$} \\
        \midrule
        Preference reranking (5 samples) & \bfseries 92.0 \\
        PreferenceFlow & 90.5 \\
        Randomized preference labels & 63.5 \\
        \bottomrule
    \end{tabular}
\end{table}

\section{Limitations and Discussion}
\label{sec:limitations}

PreferenceFlow learns local action preferences and provides no formal guarantee
of improvement in long-horizon task performance. Its supervision assumes that
a human correction is preferable to a robot chunk regenerated from the same
window-start state. Human actions are collected under closed-loop control,
whereas each robot candidate is generated from the initial observation; longer
windows may therefore yield noisier comparisons despite shared initial
conditioning. The method also relies on the base policy reaching states from
which local corrections can complete the task. Our evaluation covers four
visually guided insertion tasks on a single Franka robot. Assessing broader
applicability requires additional tasks, robots, and operators.

\section{Conclusion}
\label{sec:conclusion}

We presented \method{}, a framework that learns local action preferences from
human interventions and applies them through the QGF update to guide a frozen
flow policy. Gradient-cap and zero-gradient regularization constrain guidance
on intervention pairs and near demonstrated routine actions. On four real-world
insertion tasks, PreferenceFlow achieves \(90.5\%\) mean success, compared with
\(69\%\) for the frozen policy and \(84.5\%\) for Intervention BC. Ablations
support the proposed training objectives, independent action-token scoring,
and correctly ordered labels in the evaluated settings. These findings
support human interventions as a source of local supervision for improving
frozen generative robot policies without environment rewards.


\end{document}